\documentclass[11pt,a4paper]{article}
\usepackage[hyperref]{acl2021}
\usepackage{times}
\usepackage{adjustbox}
\usepackage{amsmath}
\usepackage{amsthm}
\usepackage{latexsym}
\usepackage{graphicx}
\usepackage{algorithm}
\usepackage{algpseudocode}
\usepackage{amssymb}
\usepackage{multirow}
\usepackage{longtable}
\usepackage{comment}
\usepackage{caption} 
\usepackage{subcaption} 
\usepackage{csquotes}

\hypersetup{
    colorlinks=true,
    linkcolor=blue,
    filecolor=magenta,      
    urlcolor=cyan,
}

\usepackage{microtype}

\aclfinalcopy 
\def\aclpaperid{2489} 

\title{Exploring the Efficacy of Automatically Generated Counterfactuals for Sentiment Analysis}

\author{
    Linyi Yang \textsuperscript{\rm1,2,3,4},
    Jiazheng Li \textsuperscript{\rm2},
    Pádraig Cunningham \textsuperscript{\rm2}, 
    Yue Zhang \textsuperscript{\rm3,4} \AND 
    Barry Smyth \textsuperscript{\rm1,2},
    Ruihai Dong \textsuperscript{\rm1,2} 
    \\
    \textsuperscript{1} The Insight Centre for Data Analytics, University College Dublin \\
    \textsuperscript{2} School of Computer Science, University College Dublin \\
    \textsuperscript{3} School of Engineering, Westlake University \\
    \textsuperscript{4} Institute of Advanced Technology, Westlake Institute for Advanced Study \\

    \texttt{\{linyi.yang, ruihai.dong, barry.smyth\}@insight-centre.org} \\
    \texttt{\{padraig.cunningham\}@ucd.ie} \\
    \texttt{\{jiazheng.li\}@ucdconnect.ie} \\
    \texttt{\{yue.zhang\}@westlake.edu.cn}
}

\date{}

\begin{document}
\maketitle

\begin{abstract}


While state-of-the-art NLP models have been achieving the excellent performance of a wide range of tasks in recent years, important questions are being raised about their robustness and their underlying sensitivity to systematic biases that may exist in their training and test data.  Such issues come to be manifest in performance problems when faced with out-of-distribution data in the field.  One recent solution has been to use counterfactually augmented datasets in order to reduce any reliance on spurious patterns that may exist in the original data. Producing high-quality augmented data can be costly and time-consuming as it usually needs to involve human feedback and crowdsourcing efforts.  In this work, we propose an alternative by describing and evaluating an approach to automatically generating counterfactual data for the purpose of data augmentation and explanation. A comprehensive evaluation on several different datasets and using a variety of state-of-the-art benchmarks demonstrate how our approach can achieve significant improvements in model performance when compared to models training on the original data and even when compared to models trained with the benefit of human-generated augmented data.

\end{abstract}

\section{Introduction}

Deep neural models have recently made remarkable advances on sentiment analysis \cite{devlin2018bert,liu2019roberta,yang2019xlnet,xie2020unsupervised}. However, their implementation in practical applications still encounters significant challenges. Of particular concern, these models tend to learn intended behavior that is often associated with spurious patterns (artifacts) \cite{jo2017measuring,slack2020fooling}. As an example, in the sentence \emph{``Nolan's films always shock people, thanks to his superb directing skills"}, the most influential word for the prediction of a positive sentiment should be \emph{``superb"} instead of \emph{``Nolan"} or \emph{``film"}. The issue of spurious patterns also partially affects the out-of-domain (OOD) generalization of the models trained on independent, identical distribution (IID) data, leading to performance decay under distribution shift \cite{quionero2009dataset,sugiyama2012machine,ovadia2019can}. 

Researchers have recently found that such concerns about model performance decay and social bias in NLP come about out-of-domain because of a sensitivity to semantically spurious signals \cite{gardner2020evaluating}, and recent studies have uncovered a problematic tendency for gender bias in sentiment analysis \cite{zmigrod2019counterfactual,maudslay2019s,lu2020gender}. To this end, one of the possible solutions is data augmentation with counterfactual examples \cite{ICLR20Counterfact} to ensure that models learn real causal associations between the input text and labels. For example, a sentiment-flipped counterfactual of last example could be \emph{``Nolan's movies always \textbf{bore} people, thanks to his \textbf{poor} directorial skills."}. When added to the original set of training data, such kinds of counterfactually augmented data (CAD) have shown their benefits on learning real causal associations and improving the model robustness in recent studies \cite{ICLR20Counterfact,ICLR21Counterfact,wang2021robustness}. Unlike gradient-based adversarial examples \cite{wang2019automatic,zhang2019generating,zang2020word}, which cannot provide a clear boundary between positive and negative instances to humans, counterfactuals could provide \emph{``human-like''} logic to show a modification to the input that makes a difference to the output classification~\cite{byrne2019counterfactuals}.

Recent attempts for generating counterfactual examples (also known as minimal pairs) rely on human-in-the-loop systems. \citet{ICLR20Counterfact} proposed a human-in-the-loop method to generate CAD by employing human annotators to generate sentiment-flipped reviews. The human labeler is asked to make minimal and faithful edits to produce counterfactual reviews. Similarly, \citet{srivastava2020robustness} presented a framework to leverage strong prior (human) knowledge to understand the possible distribution shifts for a specific machine learning task; they use human commonsense reasoning as a source of information to build a more robust model against spurious patterns. Although useful for reducing sensitivity to spurious correlations, collecting enough high-quality human annotations is costly and time-consuming. 

The theory behind the ability of CAD to improve model robustness in sentiment analysis is discussed by \citet{ICLR21Counterfact}, where researchers present a theoretical characterization of the impact of noise in causal and non-causal features on model generalization. However, methods for automatically generating CAD have received less attention. The only existing approach \cite{wang2021robustness} has been tested on the logistic regression model only, despite the fact that recent state-of-the-art methods for sentiment classification are driven by neural models. Also, their automatically generated CAD cannot produce competitive performance compared to human-generated CAD. We believe that their method does not sufficiently leverage the power of pre-trained language models and fails to generate fluent and effective CAD. In addition, the relationships between out-of-domain generalization and sensitivity to spurious patterns were not explicitly investigated by \citet{wang2021robustness}. 

To address these issues, we use four benchmark datasets (IMDB movie reviews as hold-out test while Amazon, Yelp, and Twitter datasets for out-of-domain generalization test) to further explore the efficacy of CAD for sentiment analysis. First, we conduct a systematic comparison of several different state-of-the-art models \cite{wang2021robustness}. This reveals how large Transformer-based models \cite{vaswani2017attention} with larger parameter sizes may improve the resilience of machine learning models. Specifically, we have found that for increasing parameter spaces, CAD's performance benefit tends to decrease, regardless of whether CAD is controlled manually or automatically. Second, we introduce a novel masked language model for helping improve the fluency and grammar correctness of the generated CAD. Third, we add a fine-tuned model as a discriminator for automatically evaluating the edit-distance, using data generated with minimal and fluent edits (same requirements for human annotators in \citet{ICLR20Counterfact}) to ensure the quality of generated counterfactuals. Experimental results show that it leads to significant prediction benefits using both hold-out tests and generalization tests. 

To the best of our knowledge, we are the first to automatically generate counterfactuals for use as augmented data to improve the robustness of neural classifiers, which can outperform existing, state-of-the-art, human-in-the-loop approaches. We will release our code and datasets on GitHub \footnote{https://github.com/lijiazheng99/Counterfactuals-for-Sentiment-Analysis}.

\begin{figure*}[htbp]
  \centering
  \includegraphics[width=0.95\linewidth]{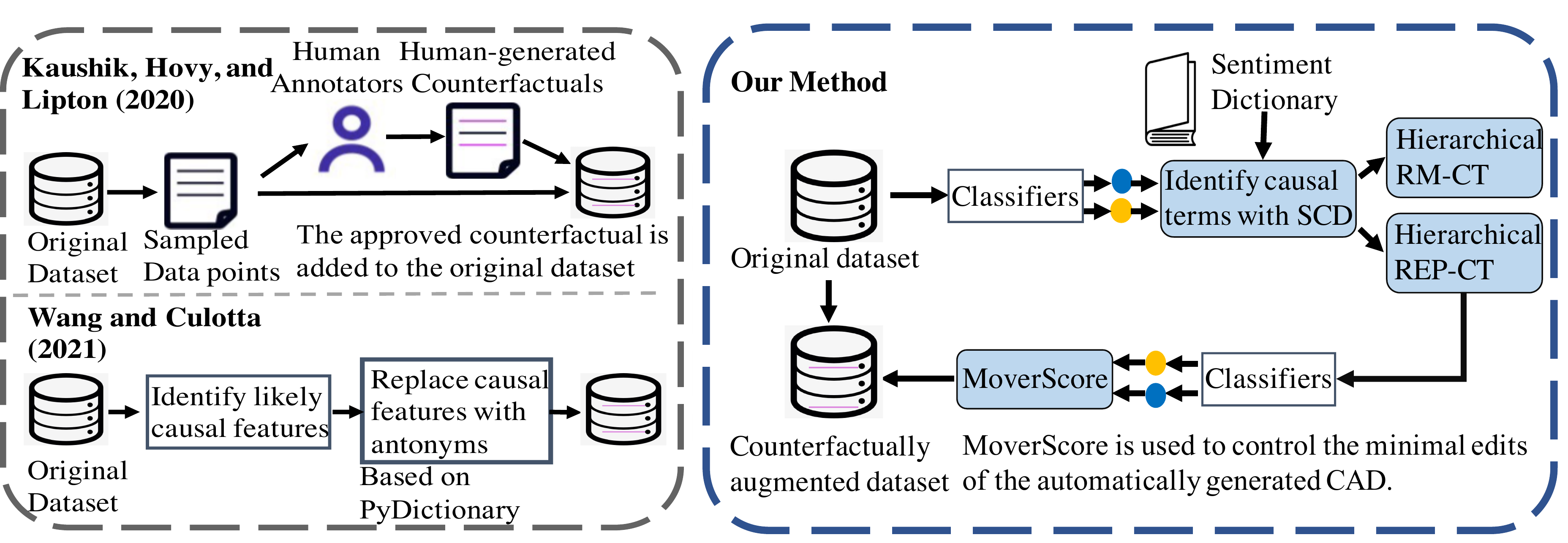}
  \caption{Overview of previous CAD methods are shown on the left side, while the pipeline of our method is shown on the right. Hierarchical RM-CT (removing the casual terms) and Hierarchical REP-CT (replacing the casual terms) are our methods for automatically generating CAD, respectively. SCD denotes \textit{sampling and sensitivity of contextual decomposition}. Sentiment Dictionary refers to the opinion lexicon published by \cite{hu2004mining}.}
\end{figure*}

\section{Related Work}
This work mainly touches on three important areas: approaches to evaluation that go beyond traditional accuracy measures \cite{bender-koller-2020-climbing,warstadt2020blimp}, the importance of counterfactuals in eXplainable AI (XAI) \cite{byrne2019counterfactuals,keane2020good}, and out-of-domain generalization in sentiment analysis \cite{kim2004determining,zhang2018sentence,zhang2019tree}.

There has been an increasing interest in the role of \textbf{Robustness Causal Thinking} in ML, often by leveraging human feedback. Recently, some of the standard benchmark datasets have been challenged \cite{gardner2020evaluating,ribeiro-etal-2020-beyond}, in which the model performance is significantly lower on contrast sets than on original test sets; a difference of up to 25\% in some cases. Researchers propose counterfactual data augmentation approaches for building robust models \cite{maudslay2019s,zmigrod2019counterfactual,lu2020gender}, and find that spurious correlations threaten the model's validity and reliability. In an attempt to address this problem, \citet{ICLR20Counterfact} explore opportunities for developing human-in-the-loop systems by using crowd-sourcing to generate counterfactual data from original data, for data augmentation. \citet{teney2020learning} shows the continuous effectiveness of CAD in computer vision (CV) and NLP. 

The idea of generating \textbf{Counterfactuals in XAI} also shares important conceptual features with our work. Since human counterfactual explanations are minimal in the sense that they select a few relevant causes \cite{byrne2019counterfactuals,keane2020good} as is the requirement of minimal edits in our generation process. This has been explored more in the field of CV \cite{goyal2019counterfactual,kenny2020generating}, but investigated less in NLP. Recent work \cite{jacovi2020aligning}  highlight explanations of a given causal format, and \citet{yang2020generating} generate counterfactuals for explaining the prediction of financial text classification since the interpretability is a more crucial requirement for the high-risk downstream applications \cite{xing2020financial,yang2020html,li2020maec}.  We propose a similar but different research question, that is, whether the automatically generated counterfactual can be used for data augmentation to build more robust models, which has not been considered by the previous methods in XAI \cite{pedreschi2019meaningful,slack2020much,yang2020semeval}.

In the case of \textbf{Sentiment Analysis}, most of the previous works report experiments using a hold-out test on the IID dataset \cite{liu2012sentiment,yang2016hierarchical,johnson2017deep}. The current state-of-the-art methods make use of large pre-trained language models (e.g., BERT \cite{devlin2018bert}, RoBERTa \cite{liu2019roberta} and SMART-RoBERTa \cite{jiang2020smart}) for calculating input represntations. It has been shown that these methods can suffer from spurious patterns \cite{ICLR20Counterfact,wang2021robustness}. Very recently, \citet{wang2021robustness} provide a starting point for exploring the efficacy of automatically generated CAD for sentiment analysis, but it is still based on IID hold-out tests only. However, spurious patterns in the training and test sets could be tightly coupled, which may limit the possibility of observing their attendant accuracy issues using a hold-out test methodology. For this reason, we designed an indirect method for evaluating the robustness of models, by comparing the performance of models trained on original and augmented data using out-of-domain data. The prediction benefit for out-of-domain data should provide some evidence about whether a model's sensitivity to spurious patterns has been successfully mitigated. The resulting counterfactuals can be used for data augmentation and can also provide contrastive explanations for classifiers, and important and desirable consideration for the recent move towards more XAI \cite{ribeiro2016should,lundberg2017unified,lipton2018mythos,pedreschi2019meaningful,slack2020much}. 



\section{Detailed Implementation}
We propose a new approach for automatically generating counterfactuals to enhance the robustness of sentiment analysis models by inverting the sentiment of causally important terms according to Algorithm \ref{alg} and based on the following stages:

\begin{enumerate}
    \item The identification of genuine causal terms using self-supervised contextual decomposition (Section 3.1).
    \item Generating counterfactual samples by (a) RM-CT (removing causal terms) and (b) REP-CT (replacing the causal terms) (Section 3.2).
    \item Selecting the human-like counterfactuals using MoverScore. \cite{zhao2019moverscore} (Section 3.3).
\end{enumerate}

The end result will be a set of counterfactuals that can be used to augment an existing dataset.


\subsection{Identifying Causal Terms}
To identify causally important terms, we propose a hierarchical method, based on the \textit{sampling and sensitivity of contextual decomposition} technique from \citet{jin2019towards}, by incrementally removing words from a sentence in order to evaluate the model's sensitivity to these words. Significant changes in model outputs suggest the removal of important terms. For example, removing the word \emph{``best''} from \emph{``The movie is the best that I have ever seen."}, is likely to alter a model's sentiment prediction more than the removal of other words from the sentence; thus \emph{``best''} is an important word with respect to this sentence's sentiment. In a similar way, phrases beginning with negative pronouns will likely be important; for instance, \emph{``not satisfy you"} is important in \emph{``This movie could not satisfy you"}.



Given a word (or phrase starting with negative limitations) \(w\) in the sentence \(s\), the importance of \(w\) can be calculated as in Equation \ref{eq:1} where \(\mathbf{s}\__{\beta} \backslash \mathbf{p}\) denotes the sentence that resulting after masking out a single word (or a negative phrase as above). We use \(l\left(\mathbf{s}\__{\beta} \backslash \mathbf{p} ; \widehat{\mathbf{s}}\right)\) to represent the model prediction after replacing the masked-out context, while \(\widehat{\mathbf{s}_{\beta}}\) is a input sequence sampled from the input \(\mathbf{s}\). \(\backslash \mathbf{p}\) indicates the operation of masking out the phrase \(p\) in a input document \(\mathcal{D}\) from the training set. The specific candidate causal terms found by this masking operation vary for different prediction models. 

\begin{equation}
\phi(\mathbf{w}, \widehat{\mathbf{s}})=\mathbb{E}_{\mathbf{s}_{\beta}}\left[\frac{l\left(\mathbf{s}\__{\beta};\widehat{\mathbf{s}_{\beta}}\right)-l\left(\mathbf{s}\__{\beta} \backslash \mathbf{p} ; \widehat{\mathbf{s}_{\beta}}\right)}{l\left(\mathbf{s}\__{\beta} ; \widehat{\mathbf{s}_{\beta}}\right)}\right]
\label{eq:1}
\end{equation}



\begin{algorithm}[t]
\caption{Generating plausible counterfactual instances.}
\textbf{Input:} \small{Test document \textbf{\(\mathcal{D}^{(n)}\)}= \(\{{P}_{1},{P}_{2},...,{P}_{n}\}\), with corresponding ground-truth labels \textbf{Y}, pre-trained Mask Language Model \textbf{MLM}, fine-tuned transformer classifier \textbf{C}, Positive Word Dictionaries \textbf{POS}, Negative Word Dictionaries \textbf{NEG}. (\textbf{pos} and \textbf{neg} are predicates for positive and negative labels)}\\
\textbf{Output:} Plausible counterfactual \textbf{\({D}^{(k)}_{cf}\)= \(\{{D}^{(k)}_{rep},{D}^{(k)}_{rm}\}\)}
\begin{algorithmic}[1]
\For {$P_k$ in $D^{(n)}$}
    \For {$S^{(i)}$, $Y_i$ in $P_{k}$}
        \State $\widehat{S}^{(i)} \gets \big\{w\in S^{(i)}~| ~(w\in POS \land Y_i=pos)~$ 
        \Statex \quad\qquad\qquad\qquad $\lor~(w\in NEG \land Y_i=neg)\big\}$
        
        \State ${S}^{(i)}_{sorted} \gets sort\big(\widehat{S}^{(i)}, key=\phi(w, \widehat{S}^{(i)})\big)(eq.1)$
        
        \State $S^{(i)}_{rm} \gets {S}^{(i)}_{sorted}[1:]$
        \State $S^{(i)}_{rep} \gets {S}^{(i)}_{sorted}$
        
        \For{$w\in~S^{(i)}_{rep}$}
            \State $W_{p} \gets MLM\big({S}^{(i)}_{mask(w)}, S^{(i)}_{rep}\big)$
            
            \State $W_c \gets \{w\in W_p~|~(w\in POS \land Y_i!=pos)~\lor~(w\in NEG \land Y_i!=neg)\big\}$
            
            \State $S^{(i)}_{rep}(w) \gets sort\big(W_{c}, key=\phi(w, W_c)\big)[0]$
        \EndFor
        \State $P^{(k)}_{rm} \gets P^{(k)}_{rm} + S^{(i)}_{rm}$
        \State $P^{(k)}_{rep} \gets P^{(k)}_{rep} + S^{(i)}_{rep}$
    \EndFor
    \State $D^{(n)}_{rm} \gets D^{(n)}_{rm} + P^{(k)}_{rm}$
    \State $D^{(n)}_{rep} \gets D^{(n)}_{rep} + P^{(k)}_{rep}$
\EndFor
\State \Return{$D^{(n)}_{rm}$, $D^{(n)}_{rep}$}
\end{algorithmic}
\label{alg}
\end{algorithm}

\subsection{Generating Human-like Counterfactuals}
This approach and the scoring function in Equation \ref{eq:1} is used in Algorithm \ref{alg} in two ways, to generate two types of plausible counterfactuals. First, it is used to identify words to \emph{remove} from a sentence to produce a plausible counterfactual. This is referred to as RM-CT and is performed by lines 3--5 in Algorithm \ref{alg}; for a sentence $S^{(i)}$, it's correctly labeled sentiment words are identified (line 3), and sorted based on Equation \ref{eq:1} (line 4) with classifier C, and the most important of these words is removed from $S^{(i)}$ to produce $S^{(i)}_{rm}$ (line 5).

Second, the REP-CT technique instead \emph{replaces} each causally important sentiment word in $S^{(i)}$ with an alternative word that has an opposing sentiment polarity (lines 6-11 in Algorithm \ref{alg}). To do this the words in $S^{(i)}$ are each considered for replacement in order of their importance (lines 6 \& 7) to create a new sentence $S^{(i)}_{rep}$. For each word $w$ we use a masked language model (MLM) to generate a set of plausible replacements, $W_p$ (line 8), and a subset of these, $W_c$, as replacement candidates if their sentiment is different from the sentiment of $S^{(i)}$, which is given by $Y_i$ (line 9). Here we are using the BERT-base-uncased as the pre-trained MLM for SVM and BiLSTM models  \footnote[1]{For Transformers-based models, we use their own pre-trained MLM (e.g., RoBERTa and XLNet) as the generator.}. The size of candidate substitutions found by MLM output is set to 100 for all models.Then, $W_c$ is sorted in descending order of importance using Equation \ref{eq:1} and the most important candidate is selected and used to replace $w$ in $S^{(i)}_{rep}$ (line 10). 

Algorithm \ref{alg} continues in this fashion to generate counterfactual sentences using RM-CT and REP-CT for each sentence in each paragraph of the target document \footnote[2]{Generating one counterfactual edit for an IMDB instance takes an average of $\approx 3.4$ seconds based on the RoBERTa-Large model.}. It returns two counterfactual documents, which correspond to documents produced from the RM-CT and REP-CT sentences; see lines 15--18.

The above approach is not guaranteed to always generate counterfactuals. Typically, reviews that cannot be transformed into plausible counterfactuals contain spurious associations that interfere with the model's predictions. For example, in our method, the negative review \emph{``The film is pretty bad, and her performance is overacted"} will be first modified as \emph{``The film is pretty good, and her performance is lifelike"}. The revised review's prediction will remain negative. Meanwhile, the word \emph{``her"} will be identified as a potential causal term. To alleviate this problem, we further conduct the substitution of synonyms for those instances that have been already modified with antonym substitution by using causal terms. As an example, we will continue replacing the word \emph{``her"} with \emph{``their"} until the prediction has been flipped; see also \citet{zmigrod2019counterfactual} for related ideas. 

In conclusion, then, the final augmented dataset that is produced of three parts: (1) counterfactuals generated by RM-CT; (2) counterfactuals generated by REP-CT; (3) adversarial examples generated by synonym substitutions.



\subsection{Ensuring Minimal Changes}
When generating plausible counterfactuals, it is desirable to make minimal changes so that the resulting counterfactual is as similar as possible to the original instance \cite{miller2019explanation,keane2020good}. To evaluate this for the approach described we use the MoverScore \cite{zhao2019moverscore} -- an edit-distance scoring metric originally designed for machine translation -- which confirms that the MoverScore for the automatic CAD instances is marginally higher when compared to human-generated counterfactuals, indicated greater similarity between counterfactuals and their original instances. The MoverScore between human-generated counterfactuals and original reviews is 0.74 on average (minimum value of 0.55) and our augmented data results in a slightly higher average score than human-generated data for all models. The generated counterfactuals and synonym substitutions that achieve a MoverScore above 0.55 are combined with the original dataset for training robust classifiers.



\section{Datasets}
Our evaluation uses three different kinds of datasets, in-domain data, challenge data, and out-of-domain data. 
\subsection{In-domain Data}
We first adopt two of the most popular benchmark datasets -- SST-2 and IMDB \cite{maas2011learning} -- to show the recent advances on sentiment analysis with the benefit of pre-trained models. However, we mainly focus on the robustness of various models for sentiment analysis in this work, rather than in-domain accuracy. Hence, following \citet{wang2021robustness} and \citet{ICLR20Counterfact}, we perform binary sentiment classification experiments on the IMDB dataset sampled from \citet{maas2011learning} that contains 1707 training, 245 validation, and 488 testing examples with challenge dataset (paired counterfactuals).

\subsection{Challenge Data}
Based on the in-domain IMDB data, \citet{ICLR20Counterfact} employ crowd workers not to \emph{label} documents, but to \emph{revise} movie review to reverse its sentiment, without making any gratuitous changes. We directly use human-generated counterfactuals by \citet{ICLR20Counterfact} as our challenge data, enforcing a 50:50 class balance. 

\subsection{Out-of-domain Data}
We also evaluate our method on different out-of-domain datasets, including Amazon reviews \cite{ni2019justifying} from six genres: beauty, fashion, appliances, gift cards, magazines, and software, a Yelp review dataset, and the Semeval-2017 Twitter dataset \cite{rosenthal2017semeval}. These have all been sampled to provide a 50:50 label split. The size of the training data has been kept the same for all methods, and the results reported are the average from five runs to facilitate a direct comparison with baselines \cite{ICLR20Counterfact,ICLR21Counterfact}.

\begin{table}[t]
\setlength{\tabcolsep}{1.33mm}
\centering
\small
\begin{tabular}{lll}

\hline
\textbf{State-of-the-art Models} & \textbf{SST-2} & \textbf{IMDB} \\ \hline
SMART-RoBERTa \cite{jiang2020smart}& \textbf{97.5} & \textbf{96.3} \\
RoBERTa-Large \cite{liu2019roberta}& 96.7 & \textbf{96.3} \\
RTC-attention \cite{zhang2019tree} & 90.3 & 88.7 \\
Bi-LSTM & 86.7 & 86.0 \\ \hline
\end{tabular}
\caption{The performance of state-of-the-art models in sentiment analysis.}
\end{table}

\begin{table*}[t]
\centering
\small
\begin{tabular}{llllllllll}
\hline
{\textbf{Models}} & \multicolumn{1}{l|}{{\textbf{Parameter}}} & \multicolumn{4}{l|}{{ \textbf{Training / Testing data}}} & \multicolumn{4}{l}{{ \textbf{AC: (Our method)}}} \\ \hline
{ } & \multicolumn{1}{l|}{{ }} & {O/O} & {CF/O} & {CF/CF} & \multicolumn{1}{l|}{{O/CF}}&{C/O}&{AC/O}&{C/CF}&{AC/CF}\\\hline
{SVM(TF-IDF)}&\multicolumn{1}{l|}{{-}}&{80.0}&{58.3}&{91.2}&\multicolumn{1}{l|}{{51.0}}&{83.7}&{\textbf{84.8}}&{87.3}&{86.1}\\
{Bi-LSTM}&\multicolumn{1}{l|}{{0.2M}}&{79.3}&{62.5}&{89.1}&\multicolumn{1}{l|}{{55.7}}&{81.5}&{\textbf{82.2}}&{92.0}&{88.5}\\\hline
\multicolumn{10}{l}{{\textbf{Transformer-based Models}}}\\\hline
{BERT [ICLR,2021]}&\multicolumn{1}{l|}{{110M}}&{87.4}&{80.4}&{90.8}&\multicolumn{1}{l|}{{82.2}}&{88.5}&{\textbf{90.6}}&{95.1}&{92.2}\\
{WWM-BERT-Large}&\multicolumn{1}{l|}{{335M}}&{91.2}&{86.9}&{96.9}&\multicolumn{1}{l|}{{93.0}}&{91.0}&{\textbf{91.8}}&{95.3}&{94.1}\\
{XLNet-Large}&\multicolumn{1}{l|}{{340M}}&{\textbf{95.3}}&{90.8}&{98.0}&\multicolumn{1}{l|}{{93.9}}&{93.9}&{\textbf{94.9}}&{96.9}&{95.5}\\
{RoBERTa-Large}&\multicolumn{1}{l|}{{355M}}&{93.4}&{91.6}&{96.9}&\multicolumn{1}{l|}{{93.0}}&{93.6}&{\textbf{94.1}}&{96.7}&{94.3}\\ \hline
\end{tabular}
\caption{The accuracy of various models for sentiment analysis using different datasets, including the human-generated counterfactual data and counterfactual samples generated by our pipeline. \textit{O} denotes the original IMDB review dataset, \textit{CF} represents the human-revised counterfactual samples, \textit{C} denotes the combined dataset consisting of original and human-revised dataset, and \textit{AC} denotes the original dataset combined with automatically generated counterfactuals. \textit{C} and \textit{AC} contain the same size of training samples (3.4K).}
\label{tab:accOM}
\end{table*}

\begin{table}[t]
\centering
\setlength{\tabcolsep}{3mm}
\small
\begin{tabular}{lll}
\hline
\textbf{Original Samples} & \textbf{Original} & \textbf{Robust} \\ \hline
\begin{tabular}[c]{@{}l@{}}Nolan's film...superb\\directing skills (\textit{POS})\\\end{tabular} & \begin{tabular}[c]{@{}l@{}} superb:$0.213$\\film:$0.446$\\Nolan:$0.028$
 \end{tabular} & \begin{tabular}[c]{@{}l@{}} $0.627$\\$0.019$\\$0.029$\end{tabular} \\ \hline
\begin{tabular}[c]{@{}l@{}}It's a poor film, but I \\must give it to the lead\\actress in this one (\textit{NEG})\end{tabular} & \begin{tabular}[c]{@{}l@{}} poor:-0.551\\film:-0.257\\actress:-0.02 \end{tabular} & \begin{tabular}[c]{@{}l@{}}-0.999\\-7e-7\\-1e-6\end{tabular} \\ \hline
\end{tabular}
\caption{Less sensitivity to spurious patterns has been shown in the robust BERT-base-uncased model.}
\end{table}

\section{Results and Discussions}
We first describe the performance of the current state-of-the-art methods on sentiment analysis based on the SST-2 and IMDB benchmark datasets. Next, we will discuss the performance benefits by using our automatically generated counterfactuals on an in-domain test. We further compare our method, human-label method, and two state-of-the-art style-transfer methods \cite{sudhakar2019transforming,madaan2020politeness} in terms of the model robustness on generalization test. Notably, we provide an ablation study lastly to discuss the influence of \emph{edit-distance} for performance benefits.

\subsection{State-of-the-art Models}

As the human-generated counterfactuals \cite{ICLR20Counterfact} are sampled from \citet{maas2011learning}, the results in Table 1 cannot be directly compared with Table 2 \footnote[3]{We can only get the human-generated counterfactual examples \cite{ICLR20Counterfact} sampled from the IMDB dataset.}. As shown in Table 1, by comparing BiLSTM to Transformer-base methods, it can be seen that remarkable advances in sentiment analysis have been achieved in recent years. On SST-2, SMART-RoBERTa \cite{jiang2020smart} outperforms Bi-LSTM by 10.8\% (97.5\% vs. 86.7\%) accuracy, where a similar improvement is observed on IMDB (96.3\% vs. 86.0\%). 

According to the results, we select the following models for our experiments, which covers a spectrum of statistical, neural and pre-trained neural methods: SVM \cite{suykens1999least}, Bi-LSTM \cite{graves2005framewise}, BERT-Base \cite{devlin2018bert}, RoBERTa-Large \cite{liu2019roberta}, and XLNet-Large \cite{yang2019xlnet}. The SVM model for sentiment analysis is from \textup{scikit-learn} and uses TF-IDF (Term Frequency-Inverse Document Frequency) scores, while the Transformer-based models are built based on the Pytorch-Transformer package \footnote[4]{\url{https://github.com/huggingface/pytorch-transformers}}. We keep the prediction models the same as \citet{ICLR20Counterfact}, except for Naive Bayes, which has been abandoned due to its high-variance performance shown in our experiments. 

In the following experiments, we only care about whether the robustness of models has been improved when training on the augmented dataset (original data \& CAD). Different counterfactual examples have been generated for different models in terms of their own causal terms in practice, while the hyper-parameters for different prediction models are all identified using a grid search conducted over the validation set.

\subsection{Comparison with Original Data}
\textbf{On the Influence of  Spurious Patterns.} As shown in Table 2, we find that the linear model (SVM) trained on the original and challenge (human-generated counterfactuals) data can achieve 80\% and 91.2\% accuracy testing on the IID hold-out data, respectively. However, the accuracy of the SVM model trained on the original set when testing on the challenge data drops dramatically (\textbf{91.2\%} vs. \textbf{51\%}), and vice versa (\textbf{80\%} vs. \textbf{58.3\%}). Similar findings were reported by \citet{ICLR20Counterfact}, where a similar pattern was observed in the Bi-LSTM model and BERT-base model. This provides further evidence supporting the idea that the spurious association in machine learning models is harmful to the performance on the challenge set for sentiment analysis.

\textbf{On the Benefits of Robust BERT.}
As shown in Table 3, we also test whether the sensitivity to spurious patterns has been eliminated in the robust BERT model. We notice that the correlations of the real causal association \emph{``superb"} and \emph{``poor"} are improved from 0.213 to 0.627 and -0.551 to -0.999, respectively. While the correlation of spurious association ``film" is decreased from 0.446 to 0.019 and -0.257 to -7e-7 on positive and the negative samples, respectively. This shows that the model trained with our CAD data does provide robustness against spurious patterns.

\textbf{On the Influence of Model Size.}
Previous works \cite{ICLR21Counterfact,wang2021robustness} have not investigated the performance benefits on larger pre-trained models. While we further conduct experiments on various Transformer-based models with different parameter sizes to explore whether the larger transformer-based models can still enjoy the performance benefits of CAD (Table \ref{tab:accOM}). We observe that although the test result can increase with the parameter size increasing (best for 94.9\% using XLNet), the performance benefits brought by human-generated CAD and the auto-generated CAD declines continuously with the parameter size increase. For example, the BERT-base-uncased model trained on the auto-generated combined dataset can receive 3.2\% (90.6\% vs. 87.4\%) improvement on accuracy while performance increases only 0.6\% (91.8\% vs. 91.2\%) on accuracy for WWM-BERT-Large. It suggests that larger pre-trained Transformer models may be less sensitive to spurious patterns.

\subsection{Comparison with Human CAD}
\textbf{Robustness in the In-domain Test.}
We can see that all of the models trained on automatic CAD -- shown as AC in the Table \ref{tab:accOM} -- can outperform the human-generated CAD varying with the models (AC/O vs. C/O) as follows: SVM (+1.1\%), Bi-LSTM (+0.7\%), BERT-base-uncased (+2.1\%), BERT-Large (+0.8\%), XLNet-Large (+1.0\%), and RoBERTa-Large (+0.5\%) when testing on the original data. If we adopt the automatic CAD (AC), we note a distinct improvement in Table \ref{tab:accOM} across all models trained on the challenge data in terms of \emph{11.3\%} in average (AC/O vs. CF/O), whereas the human-generated CAD can achieve 10.2\% accuracy improvement (C/O vs. CF/O) in average. It is noteworthy that the human-generated CAD can slightly outperform our method when testing on the human-generated (CF) data, it may be because the training and test sets of the human-generated (CF) data are generated by the same group of labelers. 

\begin{table}[t]
\centering
\small
\begin{tabular}{lll}
\hline
\textbf{Out-of-domain Test using} & & \\
\textbf{Different Training Data} & \textbf{SVM} & \textbf{BERT}  \\ \hline
\multicolumn{3}{c}{\textbf{Accuracy on Amazon Reviews}} \\ \hline
Orig \& CAD (Our Method) (3.4k) & 78.6 & \textbf{84.7} \\
Orig \& CAD (By Human) (3.4k) & 79.3 & 83.3 \\
Orig. \& \cite{sudhakar2019transforming} & 64.0 & 77.2 \\
Orig. \& \cite{madaan2020politeness} & 74.3 & 71.3 \\
Orig. (3.4k) & 74.5 & 80.0 \\ \hline
\multicolumn{3}{c}{\textbf{Accuracy on Semeval 2017 Task B (Twitter)}} \\ \hline
Orig \& CAD (Our Method) (3.4k) & 69.7 & \textbf{83.8} \\
Orig \& CAD (By Human) (3.4k) & 66.8 & 82.8 \\
Orig. \& \cite{sudhakar2019transforming} & 59.4 & 72.5 \\
Orig. \& \cite{madaan2020politeness} & 62.8 & 79.3 \\
Orig. (3.4k) & 63.1 & 72.6 \\ \hline
\multicolumn{3}{c}{\textbf{Accuracy on Yelp Reviews}} \\ \hline
Orig \& CAD (Our Method) (3.4k) & 85.5 & \textbf{87.9} \\
Orig \& CAD (By Human) (3.4k) & 85.6 & 86.6 \\
Orig. \& \cite{sudhakar2019transforming} & 69.4 & 84.5 \\
Orig. \& \cite{madaan2020politeness} & 81.3 & 78.8 \\
Orig. (3.4k) & 81.9 & 84.3 \\ \hline
\end{tabular}
\caption{Out-of-domain test accuracy of SVM and BERT-base-uncased models trained on the original (Orig.) IMDB review only, Counterfactually Augmented Data (CAD) combining with original data, and sentiment-flipped style-transfer examples.}
\end{table}

\newcommand\boldblue[1]{\textcolor{blue}{\textbf{#1}}}
\newcommand\boldred[1]{\textcolor{red}{\textbf{#1}}}

\begin{table*}[t]
\begin{center}
\small
\begin{tabular}{p{0.3\textwidth}p{0.67\textwidth}}

\hline
{ \textbf{Types of Algorithms}} & { \textbf{Examples}} \\ \hline
{ } & { Ori: Some films just simply \boldred{should not be} remade. This is one of them. In and of itself \boldred{it is not a bad film}.} \\
\multirow{-2}{*}{{ \begin{tabular}[c]{@{}l@{}}Hierarchical RM-CT:\\Remove negative limitations \end{tabular}}} & { Rev: Some films just simply \boldblue{should be} remade. This is one of them. In and of itself \boldblue{it is a bad film}.} \\ \hline
{ } & { Ori: It is \boldred{badly directed, badly acted and boring}.} \\
\multirow{-2}{*}{{ \begin{tabular}[c]{@{}l@{}}Hierarchical RE-CT:\\Replacing the causal terms\end{tabular}}} & {Rev: It is \boldblue{well directed, well acted and entertaining}.} \\ \hline
{ } & { Ori: This movie is \boldred{so bad}, it can only be compared to the all-time \boldred{worst} ``comedy": Police Academy 7. \boldred{No laughs} throughout the movie.} \\
\multirow{-2}{*}{{ \begin{tabular}[c]{@{}l@{}}Combined method:\end{tabular}}} & { Rev: This movie is \boldblue{so good}, it can only be compared to the all-time \boldblue{best} ``comedy": Police Academy 7. \boldblue{Laughs} throughout the movie.} \\ \hline
\end{tabular}
\caption{Most prominent categories of edits for flipping the sentiment performed by our algorithms, namely hierarchical RM-CT and hierarchical REP-CT.}
\end{center}
\label{tab:plain}
\end{table*}

\textbf{Robustness in the Generalization Test.}
We explore how our approach makes prediction models more robust out-of-domain in Table 4. For direct comparison between our method and the human-generated method, we adopt the fine-tuned BERT-base model trained with the augmented dataset (original \& automatically revised data). The fine-tuned model is directly tested for out-of-domain data without any adjustment. As shown in Table 4, only our method and the human-label method can outperform the BERT model trained on the original data with average 6.5\% and 5.3\% accuracy improvements, respectively. Our method also offers performance benefits over three datasets even when compared to the human-label method on BERT. 

\textbf{Neural Method vs. Statistical Method.}
As shown in Table 4, the performance of the SVM model with automatic CAD  is more robust than other automated methods \cite{sudhakar2019transforming,madaan2020politeness} across all datasets. However, the human-labeled CAD can improve Amazon reviews' accuracy compared to our method using the SVM model by 0.7\%. It indicates that human-generated data may lead to more performance benefits on a statistical model.

\subsection{Comparison with Automatic Methods}

\textbf{Automatic CAD vs. Style-transfer Methods.}
As shown in Table 4, the style-transfer results are consistent with \citet{ICLR21Counterfact}. We find that the sentiment-flipped instances generated by style-transfer methods degrade the test accuracy for all models on all kinds of datasets, whereas our method has achieved the best performance for all settings. It suggests that our method have its absolute advantage for data augmentation in sentiment analysis when compared to the state-of-the-art style-transfer models.

\textbf{Our Methods vs. Implausible CAD.}
The authors of the only existing approach for automatically generating CAD \cite{wang2021robustness} report that their methods are not able to match the performance of human-generated CAD. Our methods consistently outperform human-labeled methods on both \emph{In-domain} and \emph{Out-of-domain} tests. To further provide quantitative evidence of the influence of the \emph{edit-distance} in automatic CAD, we demonstrate an ablation study in Table 6. The result shows that the quality of the generated CAD, which is ignored in the previous work \citet{wang2021robustness}, is crucial when training the robust classifiers. In particular, the BERT model fine-tuned with implausible CAD (below the threshold) can receive comparable negative results with the style-transfer samples, alongside the performance decrease on all datasets, except for Twitter.

\begin{table}[t]
\centering
\setlength{\tabcolsep}{1.5mm}
\small
\begin{tabular}{lllll}
\hline
\textbf{Training Data} & \textbf{IMDB} & \multicolumn{3}{l}{\textbf{Out-of-domain Test}} \\
BERT-base-uncased & Orig. & Amazon & Twitter & Yelp \\ \hline
Orig. \& CAD $\uparrow$ (3.4K) & \textbf{90.6} & \textbf{84.7} & \textbf{83.8} & \textbf{87.9} \\
Orig. \& CAD $\downarrow$ (3.4K) & 87.1 & 79.5 & 73.8 & 79.0 \\
Orig. (1.7K) & 87.4 & 80.0 & 72.6 & 84.3 \\ \hline
\end{tabular}
\caption{Ablation study on the influence of the \emph{edit-distance} controlled by the threshold of MoverScore. $\uparrow$ indicates the CAD (1.7K) above the threshold, while $\downarrow$ denotes the CAD (1.7K) below the threshold.}
\end{table}

\subsection{Case Study and Limitations}

The three most popular kinds of edits are shown in Table 5. These are, negation words removal, sentiment words replacement, and the combination of these. It can be observed from these examples that we ensure the edits on original samples should be minimal and fluent as was required previously with human-annotated counterfactuals \cite{ICLR20Counterfact}. As shown in Table 5, we flipped the model's prediction by replacing the causal terms in the phrase \textit{``badly directed, badly acted and boring"} to \textit{``well directed, well acted and entertaining"}, or removing  \textit{``No laughs throughout the movie."} to \textit{``Laughs throughout the movie"} for a movie review.

We also noticed that our method may face the challenge when handling more complex reviews. For example, the sentence \textit{``Watch this only if someone has a gun to your head ... maybe."} is an apparent negative review for a human. However, our algorithm is hard to flip the sentiment of such reviews with no explicit casual terms. The technique on sarcasm and irony detection may have benefits for dealing with this challenge.



\section{Conclusion}
We proposed a new framework to automatically generate counterfactual augmented data (CAD) for enhancing the robustness of sentiment analysis models. By combining the automatically generated CAD with the original training data, we can produce more robust classifiers. We further show that our methods can achieve better performance even when compared to models trained with human-generated counterfactuals. More importantly, our evaluation based on several datasets has demonstrated that models trained on the augmented data (original \& automatic CAD) appear to be less affected by spurious patterns and generalize better to out-of-domain data. This suggests there exists a significant opportunity to explore the use of the CAD in a range of tasks (e.g., natural language inference, natural language understanding, and social bias correction.).

\section*{Impact Statement}

Although the experiments in this paper are conducted only in the sentiment classification task, this study could be a good starting point to investigate the efficacy of automatically generated CAD for building robust systems in many NLP tasks, including Natural Language Inference (NLI), Named Entity Recognition (NER), Question Answering (QA) system, etc. 

\section*{Acknowledgment}
We would like to thank Eoin Kenny and Prof. Mark Keane from Insight Centre for their helpful advice and discussion during this work. Also, we would like to thank the anonymous reviewers for their insightful comments and suggestions to help improve the paper. This publication has emanated from research conducted with the financial support of Science Foundation Ireland under Grant number 12/RC/2289\_P2 and China Postdoctoral Science Foundation under Westlake University. 

\bibliography{anthology,acl2020}

\begin{thebibliography}{58}
\expandafter\ifx\csname natexlab\endcsname\relax\def\natexlab#1{#1}\fi

\bibitem[{Bender and Koller(2020)}]{bender-koller-2020-climbing}
Emily~M. Bender and Alexander Koller. 2020.
\newblock \href {https://doi.org/10.18653/v1/2020.acl-main.463} {Climbing
  towards {NLU}: {On} meaning, form, and understanding in the age of data}.
\newblock In \emph{Proceedings of the 58th Annual Meeting of the Association
  for Computational Linguistics}, pages 5185--5198, Online. Association for
  Computational Linguistics.

\bibitem[{Byrne(2019)}]{byrne2019counterfactuals}
Ruth~MJ Byrne. 2019.
\newblock Counterfactuals in explainable artificial intelligence (xai):
  evidence from human reasoning.
\newblock In \emph{Proceedings of the 28th International Joint Conference on
  Artificial Intelligence}, pages 6276--6282. AAAI Press.

\bibitem[{Devlin et~al.(2018)Devlin, Chang, Lee, and
  Toutanova}]{devlin2018bert}
Jacob Devlin, Ming-Wei Chang, Kenton Lee, and Kristina Toutanova. 2018.
\newblock Bert: Pre-training of deep bidirectional transformers for language
  understanding.
\newblock \emph{arXiv preprint arXiv:1810.04805}.

\bibitem[{Gardner et~al.(2020)Gardner, Artzi, Basmov, Berant, Bogin, Chen,
  Dasigi, Dua, Elazar, Gottumukkala et~al.}]{gardner2020evaluating}
Matt Gardner, Yoav Artzi, Victoria Basmov, Jonathan Berant, Ben Bogin, Sihao
  Chen, Pradeep Dasigi, Dheeru Dua, Yanai Elazar, Ananth Gottumukkala, et~al.
  2020.
\newblock Evaluating models’ local decision boundaries via contrast sets.
\newblock In \emph{Proceedings of the 2020 Conference on Empirical Methods in
  Natural Language Processing: Findings}, pages 1307--1323.

\bibitem[{Goyal et~al.(2019)Goyal, Wu, Ernst, Batra, Parikh, and
  Lee}]{goyal2019counterfactual}
Yash Goyal, Ziyan Wu, Jan Ernst, Dhruv Batra, Devi Parikh, and Stefan Lee.
  2019.
\newblock Counterfactual visual explanations.
\newblock In \emph{ICML}.

\bibitem[{Graves and Schmidhuber(2005)}]{graves2005framewise}
Alex Graves and J{\"u}rgen Schmidhuber. 2005.
\newblock Framewise phoneme classification with bidirectional lstm and other
  neural network architectures.
\newblock \emph{Neural networks}, 18(5-6):602--610.

\bibitem[{Hu and Liu(2004)}]{hu2004mining}
Minqing Hu and Bing Liu. 2004.
\newblock Mining and summarizing customer reviews.
\newblock In \emph{Proceedings of the tenth ACM SIGKDD international conference
  on Knowledge discovery and data mining}, pages 168--177.

\bibitem[{Jacovi and Goldberg(2020)}]{jacovi2020aligning}
Alon Jacovi and Yoav Goldberg. 2020.
\newblock Aligning faithful interpretations with their social attribution.
\newblock \emph{arXiv preprint arXiv:2006.01067}.

\bibitem[{Jiang et~al.(2020)Jiang, He, Chen, Liu, Gao, and
  Zhao}]{jiang2020smart}
Haoming Jiang, Pengcheng He, Weizhu Chen, Xiaodong Liu, Jianfeng Gao, and Tuo
  Zhao. 2020.
\newblock \href {https://doi.org/10.18653/v1/2020.acl-main.197} {{SMART}:
  Robust and efficient fine-tuning for pre-trained natural language models
  through principled regularized optimization}.
\newblock In \emph{Proceedings of the 58th Annual Meeting of the Association
  for Computational Linguistics}, pages 2177--2190, Online. Association for
  Computational Linguistics.

\bibitem[{Jin et~al.(2019)Jin, Wei, Du, Xue, and Ren}]{jin2019towards}
Xisen Jin, Zhongyu Wei, Junyi Du, Xiangyang Xue, and Xiang Ren. 2019.
\newblock Towards hierarchical importance attribution: Explaining compositional
  semantics for neural sequence models.
\newblock In \emph{International Conference on Learning Representations}.

\bibitem[{Jo and Bengio(2017)}]{jo2017measuring}
Jason Jo and Yoshua Bengio. 2017.
\newblock Measuring the tendency of cnns to learn surface statistical
  regularities.
\newblock \emph{arXiv preprint arXiv:1711.11561}.

\bibitem[{Johnson and Zhang(2017)}]{johnson2017deep}
Rie Johnson and Tong Zhang. 2017.
\newblock Deep pyramid convolutional neural networks for text categorization.
\newblock In \emph{Proceedings of the 55th Annual Meeting of the Association
  for Computational Linguistics (Volume 1: Long Papers)}, pages 562--570.

\bibitem[{Kaushik et~al.(2020)Kaushik, Hovy, and Lipton}]{ICLR20Counterfact}
Divyansh Kaushik, Eduard Hovy, and Zachary Lipton. 2020.
\newblock \href {https://openreview.net/forum?id=Sklgs0NFvr} {Learning the
  difference that makes a difference with counterfactually-augmented data}.
\newblock In \emph{International Conference on Learning Representations}.

\bibitem[{Kaushik et~al.(2021)Kaushik, Setlur, Hovy, and
  Lipton}]{ICLR21Counterfact}
Divyansh Kaushik, Amrith Setlur, Eduard Hovy, and Zachary~C Lipton. 2021.
\newblock \href {https://openreview.net/forum?id=HHiiQKWsOcV} {Explaining the
  efficacy of counterfactually augmented data}.
\newblock In \emph{International Conference on Learning Representations}.

\bibitem[{Keane and Smyth(2020)}]{keane2020good}
Mark~T Keane and Barry Smyth. 2020.
\newblock Good counterfactuals and where to find them: A case-based technique
  for generating counterfactuals for explainable ai (xai).
\newblock In \emph{International Conference on Case-Based Reasoning (ICCBR)}.

\bibitem[{Kenny and Keane(2021)}]{kenny2020generating}
Eoin~M Kenny and Mark~T Keane. 2021.
\newblock On generating plausible counterfactual and semi-factual explanations
  for deep learning.
\newblock In \emph{AAAI}.

\bibitem[{Kim and Hovy(2004)}]{kim2004determining}
Soo-Min Kim and Eduard Hovy. 2004.
\newblock Determining the sentiment of opinions.
\newblock In \emph{COLING 2004: Proceedings of the 20th International
  Conference on Computational Linguistics}, pages 1367--1373.

\bibitem[{Li et~al.(2020)Li, Yang, Smyth, and Dong}]{li2020maec}
Jiazheng Li, Linyi Yang, Barry Smyth, and Ruihai Dong. 2020.
\newblock Maec: A multimodal aligned earnings conference call dataset for
  financial risk prediction.
\newblock In \emph{Proceedings of the 29th ACM International Conference on
  Information \& Knowledge Management}, pages 3063--3070.

\bibitem[{Lipton(2018)}]{lipton2018mythos}
Zachary~C Lipton. 2018.
\newblock The mythos of model interpretability.
\newblock \emph{Queue}, 16(3):31--57.

\bibitem[{Liu(2012)}]{liu2012sentiment}
Bing Liu. 2012.
\newblock Sentiment analysis and opinion mining.
\newblock \emph{Synthesis lectures on human language technologies},
  5(1):1--167.

\bibitem[{Liu et~al.(2019)Liu, Ott, Goyal, Du, Joshi, Chen, Levy, Lewis,
  Zettlemoyer, and Stoyanov}]{liu2019roberta}
Yinhan Liu, Myle Ott, Naman Goyal, Jingfei Du, Mandar Joshi, Danqi Chen, Omer
  Levy, Mike Lewis, Luke Zettlemoyer, and Veselin Stoyanov. 2019.
\newblock Roberta: A robustly optimized bert pretraining approach.
\newblock \emph{arXiv preprint arXiv:1907.11692}.

\bibitem[{Lu et~al.(2020)Lu, Mardziel, Wu, Amancharla, and
  Datta}]{lu2020gender}
Kaiji Lu, Piotr Mardziel, Fangjing Wu, Preetam Amancharla, and Anupam Datta.
  2020.
\newblock Gender bias in neural natural language processing.
\newblock In \emph{Logic, Language, and Security}, pages 189--202. Springer.

\bibitem[{Lundberg and Lee(2017)}]{lundberg2017unified}
Scott~M Lundberg and Su-In Lee. 2017.
\newblock A unified approach to interpreting model predictions.
\newblock In \emph{Advances in neural information processing systems}, pages
  4765--4774.

\bibitem[{Maas et~al.(2011)Maas, Daly, Pham, Huang, Ng, and
  Potts}]{maas2011learning}
Andrew~L. Maas, Raymond~E. Daly, Peter~T. Pham, Dan Huang, Andrew~Y. Ng, and
  Christopher Potts. 2011.
\newblock \href {https://www.aclweb.org/anthology/P11-1015} {Learning word
  vectors for sentiment analysis}.
\newblock In \emph{Proceedings of the 49th Annual Meeting of the Association
  for Computational Linguistics: Human Language Technologies}, pages 142--150,
  Portland, Oregon, USA. Association for Computational Linguistics.

\bibitem[{Madaan et~al.(2020)Madaan, Setlur, Parekh, Poczos, Neubig, Yang,
  Salakhutdinov, Black, and Prabhumoye}]{madaan2020politeness}
Aman Madaan, Amrith Setlur, Tanmay Parekh, Barnabas Poczos, Graham Neubig,
  Yiming Yang, Ruslan Salakhutdinov, Alan~W Black, and Shrimai Prabhumoye.
  2020.
\newblock \href {https://doi.org/10.18653/v1/2020.acl-main.169} {Politeness
  transfer: A tag and generate approach}.
\newblock In \emph{Proceedings of the 58th Annual Meeting of the Association
  for Computational Linguistics}, pages 1869--1881, Online. Association for
  Computational Linguistics.

\bibitem[{Maudslay et~al.(2019)Maudslay, Gonen, Cotterell, and
  Teufel}]{maudslay2019s}
Rowan~Hall Maudslay, Hila Gonen, Ryan Cotterell, and Simone Teufel. 2019.
\newblock It’s all in the name: Mitigating gender bias with name-based
  counterfactual data substitution.
\newblock In \emph{Proceedings of the 2019 Conference on Empirical Methods in
  Natural Language Processing and the 9th International Joint Conference on
  Natural Language Processing (EMNLP-IJCNLP)}, pages 5270--5278.

\bibitem[{Miller(2019)}]{miller2019explanation}
Tim Miller. 2019.
\newblock Explanation in artificial intelligence: Insights from the social
  sciences.
\newblock \emph{Artificial Intelligence}, 267:1--38.

\bibitem[{Ni et~al.(2019)Ni, Li, and McAuley}]{ni2019justifying}
Jianmo Ni, Jiacheng Li, and Julian McAuley. 2019.
\newblock Justifying recommendations using distantly-labeled reviews and
  fine-grained aspects.
\newblock In \emph{Proceedings of the 2019 Conference on Empirical Methods in
  Natural Language Processing and the 9th International Joint Conference on
  Natural Language Processing (EMNLP-IJCNLP)}, pages 188--197.

\bibitem[{Ovadia et~al.(2019)Ovadia, Fertig, Ren, Nado, Sculley, Nowozin,
  Dillon, Lakshminarayanan, and Snoek}]{ovadia2019can}
Yaniv Ovadia, Emily Fertig, Jie Ren, Zachary Nado, David Sculley, Sebastian
  Nowozin, Joshua Dillon, Balaji Lakshminarayanan, and Jasper Snoek. 2019.
\newblock Can you trust your model's uncertainty? evaluating predictive
  uncertainty under dataset shift.
\newblock In \emph{Advances in Neural Information Processing Systems}, pages
  13991--14002.

\bibitem[{Pedreschi et~al.(2019)Pedreschi, Giannotti, Guidotti, Monreale,
  Ruggieri, and Turini}]{pedreschi2019meaningful}
Dino Pedreschi, Fosca Giannotti, Riccardo Guidotti, Anna Monreale, Salvatore
  Ruggieri, and Franco Turini. 2019.
\newblock Meaningful explanations of black box ai decision systems.
\newblock In \emph{Proceedings of the AAAI Conference on Artificial
  Intelligence}, volume~33, pages 9780--9784.

\bibitem[{Quionero-Candela et~al.(2009)Quionero-Candela, Sugiyama,
  Schwaighofer, and Lawrence}]{quionero2009dataset}
Joaquin Quionero-Candela, Masashi Sugiyama, Anton Schwaighofer, and Neil~D
  Lawrence. 2009.
\newblock \emph{Dataset shift in machine learning}.
\newblock The MIT Press.

\bibitem[{Ribeiro et~al.(2016)Ribeiro, Singh, and Guestrin}]{ribeiro2016should}
Marco~Tulio Ribeiro, Sameer Singh, and Carlos Guestrin. 2016.
\newblock " why should i trust you?" explaining the predictions of any
  classifier.
\newblock In \emph{Proceedings of the 22nd ACM SIGKDD international conference
  on knowledge discovery and data mining}, pages 1135--1144.

\bibitem[{Ribeiro et~al.(2020)Ribeiro, Wu, Guestrin, and
  Singh}]{ribeiro-etal-2020-beyond}
Marco~Tulio Ribeiro, Tongshuang Wu, Carlos Guestrin, and Sameer Singh. 2020.
\newblock \href {https://doi.org/10.18653/v1/2020.acl-main.442} {Beyond
  accuracy: Behavioral testing of {NLP} models with {C}heck{L}ist}.
\newblock In \emph{Proceedings of the 58th Annual Meeting of the Association
  for Computational Linguistics}, pages 4902--4912, Online. Association for
  Computational Linguistics.

\bibitem[{Rosenthal et~al.(2017)Rosenthal, Farra, and
  Nakov}]{rosenthal2017semeval}
Sara Rosenthal, Noura Farra, and Preslav Nakov. 2017.
\newblock Semeval-2017 task 4: Sentiment analysis in twitter.
\newblock In \emph{Proceedings of the 11th international workshop on semantic
  evaluation (SemEval-2017)}, pages 502--518.

\bibitem[{Slack et~al.(2020{\natexlab{a}})Slack, Hilgard, Jia, Singh, and
  Lakkaraju}]{slack2020fooling}
Dylan Slack, Sophie Hilgard, Emily Jia, Sameer Singh, and Himabindu Lakkaraju.
  2020{\natexlab{a}}.
\newblock Fooling lime and shap: Adversarial attacks on post hoc explanation
  methods.
\newblock In \emph{Proceedings of the AAAI/ACM Conference on AI, Ethics, and
  Society}, pages 180--186.

\bibitem[{Slack et~al.(2020{\natexlab{b}})Slack, Hilgard, Singh, and
  Lakkaraju}]{slack2020much}
Dylan Slack, Sophie Hilgard, Sameer Singh, and Himabindu Lakkaraju.
  2020{\natexlab{b}}.
\newblock How much should i trust you? modeling uncertainty of black box
  explanations.
\newblock \emph{arXiv preprint arXiv:2008.05030}.

\bibitem[{Srivastava et~al.(2020)Srivastava, Hashimoto, and
  Liang}]{srivastava2020robustness}
Megha Srivastava, Tatsunori Hashimoto, and Percy Liang. 2020.
\newblock Robustness to spurious correlations via human annotations.
\newblock In \emph{International Conference on Machine Learning}, pages
  9109--9119. PMLR.

\bibitem[{Sudhakar et~al.(2019)Sudhakar, Upadhyay, and
  Maheswaran}]{sudhakar2019transforming}
Akhilesh Sudhakar, Bhargav Upadhyay, and Arjun Maheswaran. 2019.
\newblock “transforming” delete, retrieve, generate approach for controlled
  text style transfer.
\newblock In \emph{Proceedings of the 2019 Conference on Empirical Methods in
  Natural Language Processing and the 9th International Joint Conference on
  Natural Language Processing (EMNLP-IJCNLP)}, pages 3260--3270.

\bibitem[{Sugiyama and Kawanabe(2012)}]{sugiyama2012machine}
Masashi Sugiyama and Motoaki Kawanabe. 2012.
\newblock \emph{Machine learning in non-stationary environments: Introduction
  to covariate shift adaptation}.
\newblock MIT press.

\bibitem[{Suykens and Vandewalle(1999)}]{suykens1999least}
Johan~AK Suykens and Joos Vandewalle. 1999.
\newblock Least squares support vector machine classifiers.
\newblock \emph{Neural processing letters}, 9(3):293--300.

\bibitem[{Teney et~al.(2020)Teney, Abbasnedjad, and Hengel}]{teney2020learning}
Damien Teney, Ehsan Abbasnedjad, and Anton van~den Hengel. 2020.
\newblock Learning what makes a difference from counterfactual examples and
  gradient supervision.
\newblock \emph{arXiv preprint arXiv:2004.09034}.

\bibitem[{Vaswani et~al.(2017)Vaswani, Shazeer, Parmar, Uszkoreit, Jones,
  Gomez, Kaiser, and Polosukhin}]{vaswani2017attention}
Ashish Vaswani, Noam Shazeer, Niki Parmar, Jakob Uszkoreit, Llion Jones,
  Aidan~N Gomez, Lukasz Kaiser, and Illia Polosukhin. 2017.
\newblock Attention is all you need.
\newblock In \emph{NIPS}.

\bibitem[{Wang and Wan(2019)}]{wang2019automatic}
Ke~Wang and Xiaojun Wan. 2019.
\newblock Automatic generation of sentimental texts via mixture adversarial
  networks.
\newblock \emph{Artificial Intelligence}, 275:540--558.

\bibitem[{Wang and Culotta(2021)}]{wang2021robustness}
Zhao Wang and Aron Culotta. 2021.
\newblock Robustness to spurious correlations in text classification via
  automatically generated counterfactuals.
\newblock In \emph{AAAI}.

\bibitem[{Warstadt et~al.(2020)Warstadt, Parrish, Liu, Mohananey, Peng, Wang,
  and Bowman}]{warstadt2020blimp}
Alex Warstadt, Alicia Parrish, Haokun Liu, Anhad Mohananey, Wei Peng, Sheng-Fu
  Wang, and Samuel~R Bowman. 2020.
\newblock Blimp: The benchmark of linguistic minimal pairs for english.
\newblock \emph{Transactions of the Association for Computational Linguistics},
  8:377--392.

\bibitem[{Xie et~al.(2020)Xie, Dai, Hovy, Luong, and Le}]{xie2020unsupervised}
Qizhe Xie, Zihang Dai, Eduard Hovy, Thang Luong, and Quoc Le. 2020.
\newblock Unsupervised data augmentation for consistency training.
\newblock \emph{Advances in Neural Information Processing Systems}, 33.

\bibitem[{Xing et~al.(2020)Xing, Malandri, Zhang, and
  Cambria}]{xing2020financial}
Frank Xing, Lorenzo Malandri, Yue Zhang, and Erik Cambria. 2020.
\newblock Financial sentiment analysis: An investigation into common mistakes
  and silver bullets.
\newblock In \emph{Proceedings of the 28th International Conference on
  Computational Linguistics}, pages 978--987.

\bibitem[{Yang et~al.(2020{\natexlab{a}})Yang, Kenny, Ng, Yang, Smyth, and
  Dong}]{yang2020generating}
Linyi Yang, Eoin Kenny, Tin Lok~James Ng, Yi~Yang, Barry Smyth, and Ruihai
  Dong. 2020{\natexlab{a}}.
\newblock Generating plausible counterfactual explanations for deep
  transformers in financial text classification.
\newblock In \emph{Proceedings of the 28th International Conference on
  Computational Linguistics}, pages 6150--6160.

\bibitem[{Yang et~al.(2020{\natexlab{b}})Yang, Ng, Smyth, and
  Dong}]{yang2020html}
Linyi Yang, Tin Lok~James Ng, Barry Smyth, and Riuhai Dong. 2020{\natexlab{b}}.
\newblock Html: Hierarchical transformer-based multi-task learning for
  volatility prediction.
\newblock In \emph{Proceedings of The Web Conference 2020}, pages 441--451.

\bibitem[{Yang et~al.(2020{\natexlab{c}})Yang, Obadinma, Zhao, Zhang, Matwin,
  and Zhu}]{yang2020semeval}
Xiaoyu Yang, Stephen Obadinma, Huasha Zhao, Qiong Zhang, Stan Matwin, and
  Xiaodan Zhu. 2020{\natexlab{c}}.
\newblock \href {https://www.aclweb.org/anthology/2020.semeval-1.40}
  {{S}em{E}val-2020 task 5: Counterfactual recognition}.
\newblock In \emph{Proceedings of the Fourteenth Workshop on Semantic
  Evaluation}, pages 322--335, Barcelona (online). International Committee for
  Computational Linguistics.

\bibitem[{Yang et~al.(2019)Yang, Dai, Yang, Carbonell, Salakhutdinov, and
  Le}]{yang2019xlnet}
Zhilin Yang, Zihang Dai, Yiming Yang, Jaime Carbonell, Russ~R Salakhutdinov,
  and Quoc~V Le. 2019.
\newblock Xlnet: Generalized autoregressive pretraining for language
  understanding.
\newblock In \emph{Advances in neural information processing systems}, pages
  5753--5763.

\bibitem[{Yang et~al.(2016)Yang, Yang, Dyer, He, Smola, and
  Hovy}]{yang2016hierarchical}
Zichao Yang, Diyi Yang, Chris Dyer, Xiaodong He, Alex Smola, and Eduard Hovy.
  2016.
\newblock Hierarchical attention networks for document classification.
\newblock In \emph{Proceedings of the 2016 conference of the North American
  chapter of the association for computational linguistics: human language
  technologies}, pages 1480--1489.

\bibitem[{Zang et~al.(2020)Zang, Qi, Yang, Liu, Zhang, Liu, and
  Sun}]{zang2020word}
Yuan Zang, Fanchao Qi, Chenghao Yang, Zhiyuan Liu, Meng Zhang, Qun Liu, and
  Maosong Sun. 2020.
\newblock Word-level textual adversarial attacking as combinatorial
  optimization.
\newblock In \emph{Proceedings of the 58th Annual Meeting of the Association
  for Computational Linguistics}, pages 6066--6080.

\bibitem[{Zhang et~al.(2019)Zhang, Zhou, Miao, and Li}]{zhang2019generating}
Huangzhao Zhang, Hao Zhou, Ning Miao, and Lei Li. 2019.
\newblock Generating fluent adversarial examples for natural languages.
\newblock In \emph{Proceedings of the 57th Annual Meeting of the Association
  for Computational Linguistics}, pages 5564--5569.

\bibitem[{Zhang and Zhang(2019)}]{zhang2019tree}
Yuan Zhang and Yue Zhang. 2019.
\newblock Tree communication models for sentiment analysis.
\newblock In \emph{Proceedings of the 57th Annual Meeting of the Association
  for Computational Linguistics}, pages 3518--3527.

\bibitem[{Zhang et~al.(2018)Zhang, Liu, and Song}]{zhang2018sentence}
Yue Zhang, Qi~Liu, and Linfeng Song. 2018.
\newblock Sentence-state lstm for text representation.
\newblock In \emph{Proceedings of the 56th Annual Meeting of the Association
  for Computational Linguistics (Volume 1: Long Papers)}, pages 317--327.

\bibitem[{Zhao et~al.(2019)Zhao, Peyrard, Liu, Gao, Meyer, and
  Eger}]{zhao2019moverscore}
Wei Zhao, Maxime Peyrard, Fei Liu, Yang Gao, Christian~M Meyer, and Steffen
  Eger. 2019.
\newblock Moverscore: Text generation evaluating with contextualized embeddings
  and earth mover distance.
\newblock In \emph{Proceedings of the 2019 Conference on Empirical Methods in
  Natural Language Processing and the 9th International Joint Conference on
  Natural Language Processing (EMNLP-IJCNLP)}, pages 563--578.

\bibitem[{Zmigrod et~al.(2019)Zmigrod, Mielke, Wallach, and
  Cotterell}]{zmigrod2019counterfactual}
Ran Zmigrod, Sebastian~J Mielke, Hanna Wallach, and Ryan Cotterell. 2019.
\newblock Counterfactual data augmentation for mitigating gender stereotypes in
  languages with rich morphology.
\newblock In \emph{Proceedings of the 57th Annual Meeting of the Association
  for Computational Linguistics}, pages 1651--1661.

\end{thebibliography}
\bibliographystyle{acl_natbib}

\end{document}